\documentclass{article}

\usepackage{microtype}
\usepackage{graphicx}
\usepackage{subfigure}
\usepackage{booktabs} %

\usepackage{hyperref}

\usepackage[accepted]{icml2024}

\usepackage{amsmath}
\usepackage{amssymb}
\usepackage{mathtools}
\usepackage{amsthm}

\usepackage[capitalize,noabbrev]{cleveref}

\theoremstyle{plain}
\newtheorem{theorem}{Theorem}[section]

\newtheorem{lemma}[theorem]{Lemma}

\theoremstyle{definition}

\theoremstyle{remark}

\usepackage[textsize=tiny]{todonotes}

\graphicspath{{assets}}
\newcommand\pb[1]{\ensuremath{\left[ #1 \right]}} %
\newcommand\pc[1]{\ensuremath{\left\{ #1 \right\}}} %
\newcommand\R{\ensuremath{\mathbb{R}}} %
\newcommand\Prob{\ensuremath{\mathbb{P}}} %

\usepackage{eucal}%

\newcommand\sO{\ensuremath{\mathcal{O}}}
\newcommand\sN{\ensuremath{\mathcal{N}}}

\newcommand{\norm}[1]{\left\lVert#1\right\rVert}
\newcommand{\diag}[1]{\text{diag}\left(#1\right)}

\DeclareMathOperator*{\argmax}{\arg\,\max}
\DeclareMathOperator{\rank}{rank}
\DeclareMathOperator{\vect}{vec}
\usepackage{makecell}

\icmltitlerunning{Diversity Measurement and Subset Selection for Instruction Tuning Datasets}

\begin{document}

\twocolumn[
\icmltitle{Diversity Measurement and Subset Selection for Instruction Tuning Datasets}

\icmlsetsymbol{equal}{*}

\begin{icmlauthorlist}
\icmlauthor{Peiqi Wang}{mit}
\icmlauthor{Yikang Shen}{mitibm}
\icmlauthor{Zhen Guo}{mit}
\icmlauthor{Matthew Stallone}{mitibm}
\icmlauthor{Yoon Kim}{mit}
\icmlauthor{Polina Golland}{mit}
\icmlauthor{Rameswar Panda}{mitibm}
\end{icmlauthorlist}

\icmlaffiliation{mit}{MIT, Cambridge, MA, USA}
\icmlaffiliation{mitibm}{MIT-IBM Watson AI Lab, Cambridge, MA, USA}

\icmlcorrespondingauthor{Peiqi Wang}{wpq@mit.edu}

\icmlkeywords{Machine Learning, ICML}

\vskip 0.3in
]

\printAffiliationsAndNotice{}  %

\begin{abstract}
    We aim to select data subsets for the fine-tuning of large language models to more effectively follow instructions. Prior work has emphasized the importance of diversity in dataset curation but relied on heuristics such as the number of tasks. In this paper, we use determinantal point processes to capture the diversity and quality of instruction tuning datasets for subset selection. We propose to measure dataset diversity with log determinant distance that is the distance between the dataset of interest and a maximally diverse reference dataset. Our experiments demonstrate that the proposed diversity measure in the normalized weight gradient space is correlated with downstream instruction-following performance. Consequently, it can be used to inform when data selection is the most helpful and to analyze dataset curation strategies. We demonstrate the utility of our approach on various instruction tuning datasets.
\end{abstract}

\section{Introduction}
\label{sec:intro}

Large language models (LLMs) are powerful but unwieldy for practical use. They often require demonstrations in context to elicit proper responses and even then may generate responses not intended by users. The base language model is typically ``instruction tuned'', i.e., finetuned to predict target responses given instructions. Instruction tuning enables the base language model to perform zero-shot tasks and follow users' intent more effectively, thus improving usability. Moreover, it is an indispensable step before additional preference learning to align the language model's output to human preference \citep{ouyangTrainingLanguageModels2022}.

The number of instruction tuning datasets is rapidly growing, some with millions of data points \citep{dingEnhancingChatLanguage2023,zhengRealChat1MLargeScaleRealWorld2024}. This growth is facilitated by the ease of generating synthetic datasets by prompting LLMs \citep{wangSelfInstructAligningLanguage2023} and a growing effort to retain records of real-world user interactions with these models \citep{InTheWildChat570K2023,zhengRealChat1MLargeScaleRealWorld2024}. Finetuning on ever-increasing data demands additional computational resources. As training on low quality data (e.g., incorrect responses) can lead to suboptimal models. Some data selection or pruning is required.

Practitioners in the field face an important challenge of selecting the optimal data subset for finetuning to maximize instruction following performance subject to a fixed computational budget. While various solutions have been proposed for finding representative subsets in active learning \citep{senerActiveLearningConvolutional2018}, their applicability to natural language datasets remains underexplored. For instance, active learning methods that search for subsets with diverse weight gradients \citep{ashDeepBatchActive2019} were ineffective in our initial studies as they prioritized data points with short responses or those with large weight gradient norms. Most related methods aim to provide sufficient coverage of instruction tuning examples in the space of decoder-based language models' output token embeddings \citep{bukharinDataDiversityMatters2023,liuWhatMakesGood2024} that lacks semantic structure \citep{leDistributedRepresentationsSentences2014}. Moreover, ensuring diversity in the embedding space of encoder-based masked language models is limited by encoders' short context length.

Practitioners also grapple with a closely related question of estimating how much data allocated for model finetuning would achieve comparable performance with that of the entire dataset. One approach involves assigning a score to each dataset that indicates the extent to which a dataset can be reduced without compromising performance after model finetuning. While various scoring methods exist, here we focus on dataset diversity. Common measures of dataset diversity often rely on intuitive heuristics, e.g., the number of tasks \citep{weiFinetunedLanguageModels2022,sanhMultitaskPromptedTraining2022}, topics and user intents \citep{luInsTagInstructionTagging2024}, or do not scale well with the dataset size \citep{friedmanVendiScoreDiversity2023}.

We turn to determinantal point processes (DPPs) \citep{kuleszaDeterminantalPointProcesses2012} to identify diverse subsets of high quality instruction tuning data. We investigate several choices of data representations that capture data points' similarity and find that the radial basis kernel applied to the \textit{normalized} weight gradients of the model is particularly effective when selecting from datasets that are less diverse.

In addition, we measure dataset diversity with \textit{log determinant distance} that is the difference between the log determinant of kernel matrix of a maximally diverse dataset and that of the dataset under consideration, normalized by the dataset size. Log determinant distance is readily computable from the MAP inference algorithm that identifies the optimal subset. We demonstrate that log determinant distance is correlated with instruction following performance when using weight gradients as the data representation. As a result, the diversity measure can be used to evaluate the utility of instruction tuning datasets for finetuning and to predict, before any finetuning takes place, the extent to which we can prune data without sacrificing model performance. In addition, we investigate the implications of curation strategies on dataset diversity.

\section{Related Work}
\label{sec:related_work}

\subsection{Instruction Tuning Datasets}
\label{sec:instruction_tuning_dataset}

Diversity and quality are recurring themes in the curation of instruction tuning datasets. Early instruction tuning datasets, e.g., Super-NaturalInstructions \citep{wangSuperNaturalInstructionsGeneralizationDeclarative2022} and FLAN \citep{weiFinetunedLanguageModels2022,chungScalingInstructionFinetunedLanguage2022}, are adapted from existing natural language processing benchmarks, with a particular focus on scaling the number of tasks and incorporating a variety of prompt templates to encourage task generalization and robustness to prompt wordings.

Some instruction tuning datasets are curated using Self-Instruct and its variants \citep{honovichUnnaturalInstructionsTuning2023,wangSelfInstructAligningLanguage2023} that prompt a LLM to generate a wide array of instructions and high-quality responses. These datasets, e.g., Alpaca \citep{alpaca2023}, are typically distilled from performant language models that underwent finetuning to generate user-preferred responses, e.g., variants of InstructGPT \citep{ouyangTrainingLanguageModels2022}, and are well-suited for the purposes of creating a chat assistant. Moreover, they are distilled from increasingly powerful LLMs, e.g., GPT4-Alpaca \citep{pengInstructionTuningGPT42023}, and contain more complex instructions, e.g., WizardLM \citep{xuWizardLMEmpoweringLarge2024}, step-by-step explanations in the responses, e.g., Orca \citep{mukherjeeOrcaProgressiveLearning2023}, or multi-turn conversations, e.g., UltraChat \citep{dingEnhancingChatLanguage2023}.

Another family of instruction tuning datasets aims to better reflect LLMs' real-world use cases that include significant human authorship. Some are manually curated from sources with helpful responses such as Reddit, e.g., LIMA \citep{zhouLIMALessMore2023}, or from company employees, e.g., Dolly \citep{DatabricksBlog2023DollyV2}. Alternatively, real-world user interactions are curated with state-of-the-art LLMs from the internet, e.g., ShareGPT, RealChat-1M \citep{zhengRealChat1MLargeScaleRealWorld2024}, WildChat \citep{InTheWildChat570K2023}. These datasets cover a wide range of topics and user intents, capturing real-world use scenarios.

Here we systematically study the relative diversity of aforementioned datasets and its impact on instruction following performance. Our experiments yield insights into the efficacy of the different curation approaches, e.g., distillation and manual annotation.

\subsection{Data Selection}
\label{sec:data_selection}

Analogous to data selection for finetuning, active learning selects informative examples to label from a pool of unlabeled examples subject to a fixed labeling budget. Our approach is closely related to research that formulates active learning as core-set selection, i.e., finding the representative data subset \citep{tsangCoreVectorMachines2005,wellingHerdingDynamicalWeights2009}. Examples include searching for a covering of the full dataset with the smallest cover radius by solving the k-center problem \citep{senerActiveLearningConvolutional2018} and identifying subsets that are sufficiently spread out using \texttt{k-means++} initialization \citep{ashDeepBatchActive2019}. Related, data pruning methods remove redundant data points that are too close to each other \citep{abbasSemDeDupDataefficientLearning2023} or to their respective cluster centroids \citep{sorscherNeuralScalingLaws2022}. Similarly, our work uses DPPs to model data subsets and relies on a greedy MAP algorithm \citep{chenFastGreedyMAP2018} to identify diverse subsets. The choice of distance metric and data representations is crucial. Prior works have employed the $\ell$-2 distance between neural network activations \citep{senerActiveLearningConvolutional2018,sorscherNeuralScalingLaws2022,abbasSemDeDupDataefficientLearning2023} or between weight gradients of the log likelihood \citep{huangActiveLearningSpeech2016,ashDeepBatchActive2019}. Here we investigate several data similarity measures on instruction tuning datasets.

While choosing diverse subsets is driven by the notion that similar data points are redundant, an alternative approach is motivated by the assumption that certain data points provide more value than others. Specifically, many data selection algorithms define a quality score for each data point and select the portion of the dataset with the highest scores. Various quality scoring functions have been proposed for classification tasks, including the norm of the weight gradient \citep{settlesActiveLearningLiterature2009,huangActiveLearningSpeech2016,paulDeepLearningData2021}, the number of times an example transitions from correctly classified to misclassified (i.e., ``forgotten'') during training \citep{tonevaEmpiricalStudyExample2019}, the variability of the ground-truth label likelihood over the course of training \citep{swayamdiptaDatasetCartographyMapping2020}, and the average $\ell$-2 norm of the classification error vector \citep{paulDeepLearningData2021}. Rather than propose new scoring functions, we focus on evaluating the efficacy of existing quality scores for instruction tuning subset selection. Our approach of modeling data subsets with DPPs accommodates arbitrary scores and aims to strike a balance between choosing data points with high quality scores and ensuring diversity within the selected subset.

\subsection{Selecting Natural Language Data}

Our work is adjacent to research that selects datasets for pretraining LLMs. Unfiltered pretraining text corpora like Common Crawl \citep{commoncrawl2010} are not ideal because they contain a large number of unintelligible documents and some documents are repeated many times. To discard low-quality documents, past work has relied on quality scores that scale well with the size of the dataset, e.g., ratings from Reddit users \citep{brownLanguageModelsAre2020} or the perplexity of the document computed by a pretrained language model \citep{wenzekCCNetExtractingHigh2020,marionWhenLessMore2023}. Approximate string matching algorithms, e.g., MinHash \citep{broderResemblanceContainmentDocuments1998}, are employed to detect pairs of documents with high n-gram overlap \citep{leeDeduplicatingTrainingData2022}. In contrast, properly curated instruction tuning datasets usually contain well-written texts with few repeated examples. Therefore, basic quality filters are less important. The challenge is to find informative axes of variations important for instruction following performance, e.g., topics, tasks, and user intents, which is the focus of our work.

One might want to select instruction tuning datasets for computing efficiency. Many use quality scores to rank and select data points including simple natural language indicators like coherence \citep{caoInstructionMiningHighQuality2023} or perplexity \citep{liQuantityQualityBoosting2023}, and the LLM's rating of data points based on metrics such as helpfulness \citep{chenAlpaGasusTrainingBetter2024,liuWhatMakesGood2024}. Others select data subsets with sufficient coverage of topics and user intents \citep{luInsTagInstructionTagging2024}. Our approach is closely related to methods that balance quality and diversity, e.g., by solving a variant of the facility location problem \citep{bukharinDataDiversityMatters2023} or prioritize high quality data points while avoiding duplicates \citep{liuWhatMakesGood2024}. In contrast, we model data subsets with DPPs that naturally emit a diversity metric over datasets that correlates well with the downstream instruction following performance. This metric is useful for predicting improvements in the instruction following performance and for comparing the diversity of instruction tuning datasets.

\section{Method}
\label{sec:method}

\subsection{Subset Selection with DPPs}

A point process on a set of $N$ items is a probability distribution over all subsets of $\pb{N}$. A DPP~$P$ is a point process where the probability measure is parameterized by a positive semi-definite matrix $L\in \R^{N\times N}$, i.e., $P(Y) \propto \det(L_Y)$ for any subset $Y\subset \pb{N}$. $L_Y \equiv \pb{L_{ij}}_{i,j\in Y}$ is a sub-matrix of $L$ indexed by $Y$ in rows and columns. Intuitively, the diagonal elements of $L$ are related to the marginal probability of including the particular items, i.e., $P(\pc{i})\propto L_{ii}$. The off-diagonal elements of $L$ represents the similarity between items. Similar items are less likely to co-occur, i.e., $P(\pc{i,j})\propto L_{ii}L_{jj}-L_{ij}L_{ji}$. 

Any positive semi-definite matrix $L$ can be expressed as a Gram matrix $VV^T $ for some matrix $V\in\R^{N\times D}$. Each row of $V$ can be viewed as a feature vector for $i$-th item. The absolute value of the determinant of $L_Y$ is the volume of the parallelepiped spanned by rows of $V$. Therefore, a high probability subset under $V$ is a subset whose feature vectors span a large volume \citep{kuleszaDeterminantalPointProcesses2012}.

Given a dataset with $N$ items $\pc{x_n}_{n=1}^N$, we parameterize a determinantal point process $P$ with a kernel matrix $K\in\R^{N\times N}$ that measures the similarity between data points and possibly a vector $q \in \R^N$ that indicates the quality of each data point. For instance, we can treat the cosine similarity between language models' output token embeddings as the similarity measure and the perplexity of the response conditioned on the instruction as data quality.

To select a moderately large subset of size $M$, the inner product kernel on features of dimension~$D\ll M$ is unsuitable due to rank deficiency. Specifically, any subset $Y$ with $|Y| > \rank(L)$ has zero probability mass $P(Y) \propto \det(L_Y) = 0$ and therefore the size of the most likely subset under~$P$ is upper bounded by~$\rank(L)$. Instead, we use kernel functions that induce full rank Gram matrices, e.g., the radial basis function (RBF) kernel $ K_{ij} = \exp\{- \gamma \norm{x_i-x_j}^2\}$, where a larger value of~$\gamma$ implies that the repulsive force between data points is more local. For data representations that are normalized to unit length, the radial basis function kernel reduces to $K_{ij} = \exp\pc{2\gamma x_i^Tx_j}$.

Following \citet{kuleszaStructuredDeterminantalPoint2010}, we define $L_{ij} = K_{ij} q_i q_j$. This is equivalent to scaling the kernel feature map by a scalar quality score. As long as $K$ is positive semi-definite, so is $L$. 
This structure enables us to model similarity and quality independently while considering both components during inference. Moreover, the probability of any subset $Y\subset \pb{N}$ factors, i.e.,
\begin{equation*}
    \log P(Y)
        \propto \sum_{i\in Y} \log q_i^2 + \log\det(K_Y).
\end{equation*}
The log likelihood is maximized for subsets with high quality (1st term) and diversity (2nd term). Similar to \citet{chenFastGreedyMAP2018}, we introduce a hyperparameter $\lambda\in[0,1]$ to control the relative importance of diversity and quality:
\begin{equation}
    \label{eq:log_prob_dpp_weighted_by_lambda}
    \log P(Y)
        \propto \lambda  \sum_{i\in Y} \log q_i + (1-\lambda) \log\det(K_Y),
\end{equation}
that corresponds to a DPP parameterized by the kernel matrix $L = \diag{e^{\beta q}} K \diag{e^{\beta q}}$ with $\beta = \lambda / (2(1-\lambda))$.

\begin{figure*}[ht]
    \vskip -0.05in
    \begin{center}
        \centerline{\includegraphics[width=\textwidth]{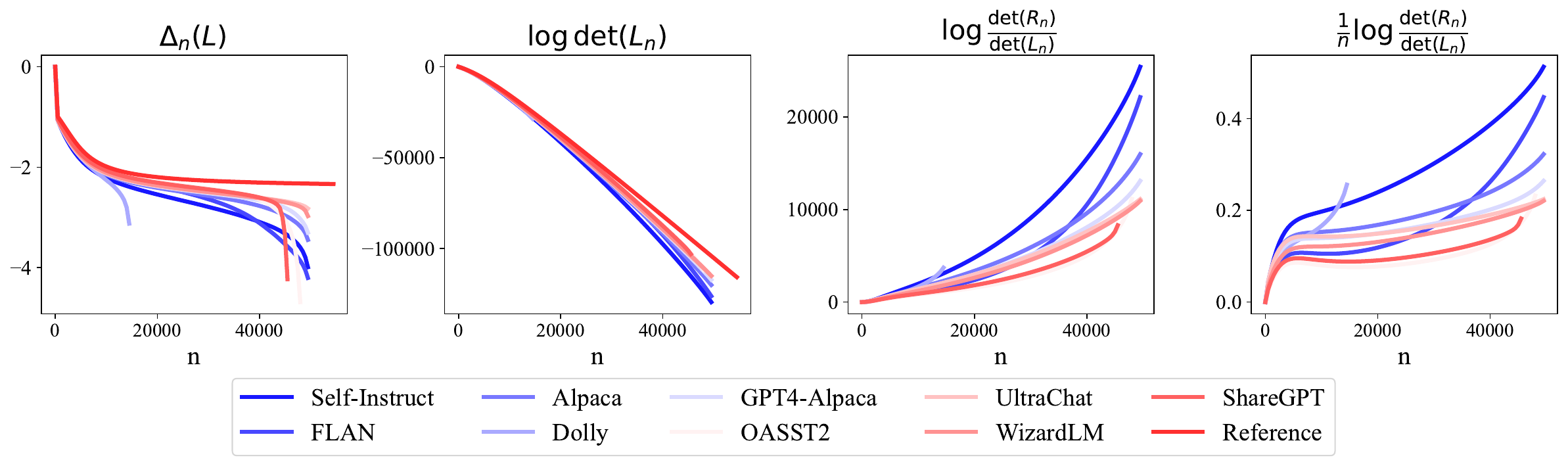}}
        \vskip -0.05in
        \caption{Step-by-step demonstration to compute the log determinant distance on a set of instruction tuning datasets of varying diversity. The marginal gain curve $\Delta_n(L)$ is derived from the greedy MAP algorithm for DPPs (1st figure). $\log\det(L_n)$ is the cumulative marginal gain curve (2nd figure). Note, scaling the kernel matrix $L$ by a constant $c>0$ shifts $\log\det(L_n)$ linearly by $n\log(c)$, complicating result interpretation. Moreover, $\log\det(L_n)$ is heavily influenced by dataset size, e.g., Dolly contains 15k examples and has a much larger $\log\det(L)$ compared to that of the other datasets subsampled to include roughly 50k examples despite it having a comparatively smaller $\log\det(L_{\text{15k}})$. To address these challenges, we compute the difference between the log determinant for a maximally diverse ``reference'' dataset and each dataset of interest (3rd figure) and then divide by dataset size (4th figure). The log determinant distance for a dataset is the value of the corresponding curve $(1/n)\log(\det(R_n)/\det(L_n))$ at the last iteration.}
        \label{fig:fig_demo_how_to_compute_diversity_LM_Weight_Gradient}
    \end{center}
    \vskip -0.25in
\end{figure*}

Given a data budget $M$, we pose subset selection as maximum a posteriori (MAP) inference under distribution $P$ with a cardinality constraint:
\begin{equation}
    \label{eq:map_inference}
    Y^*
        = \argmax_{Y\subset \pb{N}:\; |Y|=M} \det(L_Y).
\end{equation}
Although this problem is NP-hard \citep{koExactAlgorithmMaximum1995}, the log probability in \cref{eq:log_prob_dpp_weighted_by_lambda} is submodular \citep{gillenwaterNearOptimalMAPInference2012} and therefore can be solved efficiently with a greedy algorithm \citep{nemhauserAnalysisApproximationsMaximizing1978}. We use \citet{chenFastGreedyMAP2018}'s implementation with $\mathcal{O}(\text{NMD})$ time and $\mathcal{O}(\text{N(M+D)})$ memory complexity, respectively. The costly evaluation of kernel matrix entries at each iteration can be parallelized on the GPU at the memory cost of $\mathcal{O}(\text{ND})$. The algorithm is a feasible solution for selecting instruction tuning datasets at the current scale of several hundred thousand examples.

The greedy MAP inference algorithm \citep{chenFastGreedyMAP2018} grows the set of indices $S_1 \subset \cdots, S_N \subset \pb{N}$ by adding
\begin{align*}
    i^*(S)
        &= \argmax_{i\in \pb{N}\setminus S} \pb{\log\det( L_{S\cup \pc{i}} ) - \log\det( L_{S} )}
\end{align*}
to the set at each iteration. We define $L_{n} \triangleq L_{S_n}$ as shorthand for the kernel matrix $L$ indexed by the greedy solution $S_n$ at the $n$-th iteration ($L_N \equiv L$). The marginal gains $\Delta_1(L)=\log\det(L_1)$ and
\begin{align*}
    \Delta_n(L)
        &= \log\det(L_{n}) - \log\det(L_{n-1}) \\ 
        &= \log\frac{\det(L_{n})}{\det(L_{n-1})}, \quad n=2,3,\cdots
\end{align*}
approximate the rate of change in diversity of selected subsets $\pc{S_n}$ over the iterations. Larger marginal gains means the selected item contributes more to the diversity of the already selected subset. The unnormalized probability for the whole dataset is the sum of the marginal gains, i.e.,
\begin{align*}
    \log\det(L)
        = \sum_{n=1}^N \Delta_n(L).
\end{align*}

\subsection{Log Determinant Distance as a Measure of Diversity}

We propose a novel way to measure dataset diversity that is a byproduct of solving the MAP inference problem in \cref{eq:map_inference}. The measure of diversity depends entirely on the kernel that defines the DPP. For a fixed kernel function, we can compare dataset diversity quantitatively.

While $\log\det(L)$ may seem like a natural choice, it is unsuitable for measuring dataset diversity for two reasons. First, $\log\det(L)$ is not invariant to scaling of the kernel matrix, leading to widely different values that complicate the interpretation of results. For instance, if a kernel matrix is scaled by a constant $c>0$, $\log\det(cL) = N\log(c) + \log\det(L)$ changes by $N\log(c)$ for the same dataset. Second, $\log\det(L)$ depends heavily on the dataset size, particularly when there are significant marginal gains for each item selected. \cref{fig:fig_demo_how_to_compute_diversity_LM_Weight_Gradient} illustrates these problems.

To address the aforementioned challenges, we introduce a reference dataset that is maximally diverse for comparison. For example, we can generate the reference dataset by sampling at random on the hypersphere to ensure maximum diversity. We use $R$ to denote the reference dataset's kernel matrix computed using the same kernel function $k(\cdot,\cdot)$. We define \textit{Log Determinant Distance} as
\begin{align}
    \label{eq:ldd_definition}
    \text{LDD}
        &\triangleq \frac{1}{N} \log\frac{\det(R)}{\det(L)}.
\end{align}
The log determinant distance measures the average deviation of the volume of the parallelepiped spanned by the rows of the Gram matrix decomposition of $L$, i.e., $|\det(L)|$, from the largest possible volume, e.g., $|\det(R)|$. A smaller log determinant distance implies that the dataset is closer to the maximally diverse reference dataset, and therefore is more diverse. Alternatively, we can interpret the log determinant distance as the deficit in the average contribution of a data point to dataset diversity from optimum:
\begin{align*}
    \text{LDD}
        &\equiv \frac{1}{N} \sum_{n=1}^N \left( \Delta_n(R) - \Delta_n(L) \right).
\end{align*}
The log determinant distance can be readily computed from the determinants of the kernel matrices $\det(L)$ and $\det(R)$ obtained by running the greedy MAP algorithm \citep{chenFastGreedyMAP2018} twice. \cref{fig:fig_demo_how_to_compute_diversity_LM_Weight_Gradient} illustrates a step-by-step computation of the log determinant distance from marginal gains.

Given our assumption that the reference dataset is maximally diverse, i.e., $|\det(R)|\geq |\det(L)|$ for any kernel matrix $L$, the non-negativity property holds: $\text{LDD} \geq 0$. Moreover, it is straightforward to show that the log determinant distance is invariant to scaling of kernels. The log determinant distance is also invariant to permutation of datasets since it is based on matrix determinants. In summary, the log determinant distance possesses favorable properties for measuring the dataset diversity.

\subsection{Weight Gradient as Data Representation}

We use the language model's weight gradient $\nabla_{\theta} \ell(x; \theta)$ of scalar-valued loss function $\ell$ as the data representation for data point $x$. As an example, $\ell$ can be the average log likelihood of the response conditioned on the instructions. For LLMs, the full weight gradient consists of billions of elements, rendering kernel computation infeasible. In this work, we apply two Johnson-Lindenstrauss (JL) transforms \citep{johnsonExtensionsLipschitzMappings1984} consecutively on weight gradients to reduce their dimensionality.

We first apply JL transform implicitly via Low-Rank Adaptation (LoRA) \citep{huLoRALowRankAdaptation2022} to reduce memory as well as computation since most derivatives are neither stored nor computed. For weight matrix $W\in \R^{m\times n}$ in a fully connected layer, LoRA enforces a rank $r \ll min(m,n)$ update to the weight matrix that is a composition of two matrices: $B\in\R^{m\times r}$ and $A\in\R^{r\times n}$. For input activation $z\in\R^n$, the output activation $h\in\R^m$ after the update is
\begin{align*}
    h = (W + \Delta W)z = Wz + BAz.
\end{align*}
We initialize $A$ to $\sN(0, r^{-1})$ to construct a distance preserving random projection matrix and $B$ to zero to preserve the forward pass activations.
To obtain a lower-dimensional representation of the full weight gradient $\nabla_W\ell$, we use LoRA at initialization to compute 
\begin{align}
    \label{eq:grad_B_is_projection_of_grad_W}
    \nabla_{B} \ell
        &= \nabla_{h} \ell \cdot z^T A^T = \nabla_{W}\ell \cdot A^T.
\end{align}
Here, we use $\ell \equiv \ell(x;\theta)$ for brevity. We also explored approaches that use LoRA to project $\nabla_W \ell$ onto a vector of size $r$, instead of $m$ vectors of size $r$. For example, we can sum over the rows of $\nabla_B \ell$ or rows of $\nabla_B \ell$ after shifting the $i$-th row by $i$ positions. We found these approaches induce a larger pairwise distance error.

Note that $A$ is not applied to the entire weight gradient $\nabla_W\ell$. Instead, each row in $\nabla_W \ell$ of dimension $n$ is projected to the corresponding row in $\nabla_B \ell$ of dimension $r$. The typical Johnson-Lindenstrauss Lemma also holds in this case.
\begin{lemma}
    \label{lemma:jl_lemma_rowwise_projection}
    Let $\epsilon,\delta>0$. If $r = \sO(\log(1/\delta)/\epsilon^2)$, then
    \begin{align*}
        \left|
            \norm{\vect(\nabla_B \ell)}_2^2 - \norm{\vect(\nabla_W\ell)}_2^2
        \right|
            \leq \epsilon
    \end{align*}
    with probability at least $1-\delta$.
\end{lemma}
The proof in \cref{sec:appendixA_jl_lemma_rowwise_projection_proof} involves simple application of the union bound.

We then apply the sparse JL transform to the concatenation of $\vect(\nabla_B \ell)$ for every fully connected layer in the neural network to further reduce storage and compute cost. Using a sparse projection matrix is necessary since concatenated $\vect(\nabla_B \ell)$ is still too costly to work with as it contains $mlr$ entries where $l$ is the number of fully connected layer in the network.

\cref{lemma:jl_lemma_rowwise_projection} can be extended trivially to include the second JL transform. It immediately follows that the two JL transforms together preserve the pairwise distance between weight gradients, in the same way that a single JL transform does on the entire weight gradient.

\section{Experiments}
\label{sec:experiments}

\subsection{Implementation Details}
\label{sec:implementation_details}

\paragraph{Dataset} 
We employ a collection of instruction tuning datasets to understand the effect of data on model's intruction following performance and to evaluate their relative diversity: FLAN \citep{weiFinetunedLanguageModels2022}, Self-Instruct \citep{wangSelfInstructAligningLanguage2023}, Dolly \citep{DatabricksBlog2023DollyV2}, Alpaca \citep{alpaca2023}, GPT-4Alpaca \citep{pengInstructionTuningGPT42023}, OASST2 \citep{kopfOpenAssistantConversationsDemocratizing2023}, Orca \citep{mukherjeeOrcaProgressiveLearning2023}, UltraChat \citep{dingEnhancingChatLanguage2023}, WizardLM \citep{xuWizardLMEmpoweringLarge2024}, and ShareGPT. We also evaluate the diversity of preference datasets: OpenAI-Summarization \citep{stiennonLearningSummarizeHuman2020}, SHP \citep{ethayarajhUnderstandingDatasetDifficulty2022a}, UltraFeedback \citep{cuiUltraFeedbackBoostingLanguage2024}, and HH-RLHF \citep{baiTrainingHelpfulHarmless2022}. For each dataset, we remove examples with sequence lengths greater than 2,048 to ensure the language model learns to generate the end-of-sequence token properly. Except for Dolly and OASST2 that contain fewer examples, the aforementioned datasets are subsampled to 50,000 examples to control for the effect of dataset size.

\paragraph{Model \& Training} 
In all experiments, we finetune Llama-7b \citep{touvronLLaMAOpenEfficient2023} for 3 epochs with learning rate of 2e-5 and a batch size of 128. We use AdamW optimizer with no weight decay and linearly decay the learning rate after warmup for 3\% of the total number of training steps.

\paragraph{Evaluations} 

We evaluate the performance of instruction-following models using a few benchmarks that measure distinct aspects of the model: factual knowledge across various subjects like engineering and law with Massive Multitask Language Understanding (MMLU) \citep{hendrycksMeasuringMassiveMultitask2020}, reasoning on math problems using Grade School Math (GSM) \citep{cobbeTrainingVerifiersSolve2021} and on general reasoning problems with the Big-Bench Hard benchmark (BBH) \citep{suzgunChallengingBIGBenchTasks2022}, multilinguality with TydiQA \citep{clarkTyDiQABenchmark2020}, and coding skills with Codex-Eval \citep{chenEvaluatingLargeLanguage2021}. We use \textsc{Benchmarks Avg} to denote the average performance across all aforementioned benchmarks. We use Alpaca-Eval \citep{duboisAlpacaFarmSimulationFramework2023} to evaluate instruction following. Specifically, We use \textsc{AlpacaEval \% Win} to denote the proportion of times a model's generation is preferred by GPT-4 over davinci-003's response and \textsc{AlpacaEval Len} as the average number of tokens in a model's responses. We follow the evaluation procedure in \citep{wangHowFarCan2023} closely to enable fair comparisons.

\paragraph{Kernel Function}
To compute the kernel matrix $L$, we fix the kernel function to radial basis kernel and vary the data representations. We employ Llama-7b representing decoder-only language model to compute the average output token embeddings (\textsc{Llama Emb}) \& the weight gradients vectors (\textsc{Llama $\nabla_{\theta}\ell$}), and MPNet \citep{songMPNetMaskedPermuted2020} representing  encoder-only masked language model to compute the average output token embeddings of instructions (\textsc{Mpnet Emb}). We normalize these data representations to unit length and use the abbreviation $\textsc{Not Norm.}$ to imply unnormalized vectors. We set $\gamma=1$ for both \textsc{Llama $\nabla_{\theta}\ell$} and \textsc{MpNet Emb}, $\gamma=10$ for \textsc{Llama Emb}, and $\gamma=0.01$ for \textsc{Llama $\nabla_{\theta}\ell$ Not Norm.}. The choice of $\gamma$ is not critical, as long as the greedy MAP inference algorithm does not terminate prematurely and that the kernel values do not underflow.

\paragraph{Log Determinant Distance}
To compute the log determinant distance of a dataset, we generate a reference dataset by sampling vectors randomly on the surface of a $D$ dimensional hypersphere; $D=4096$ for \textsc{Llama Emb} and \textsc{Llama $\nabla_{\theta}\ell$}, and $D=768$ for \textsc{Mpnet Emb}. We then use the greedy MAP algorithm \citep{chenFastGreedyMAP2018} to obtain the determinants of the kernel matrices $\det(L)$ and $\det(R)$, from which we compute the log determinant distance in \cref{eq:ldd_definition}.

\begin{figure}[ht]
    \vskip 0.1in
    \begin{center}
    \centerline{\includegraphics[width=\columnwidth]{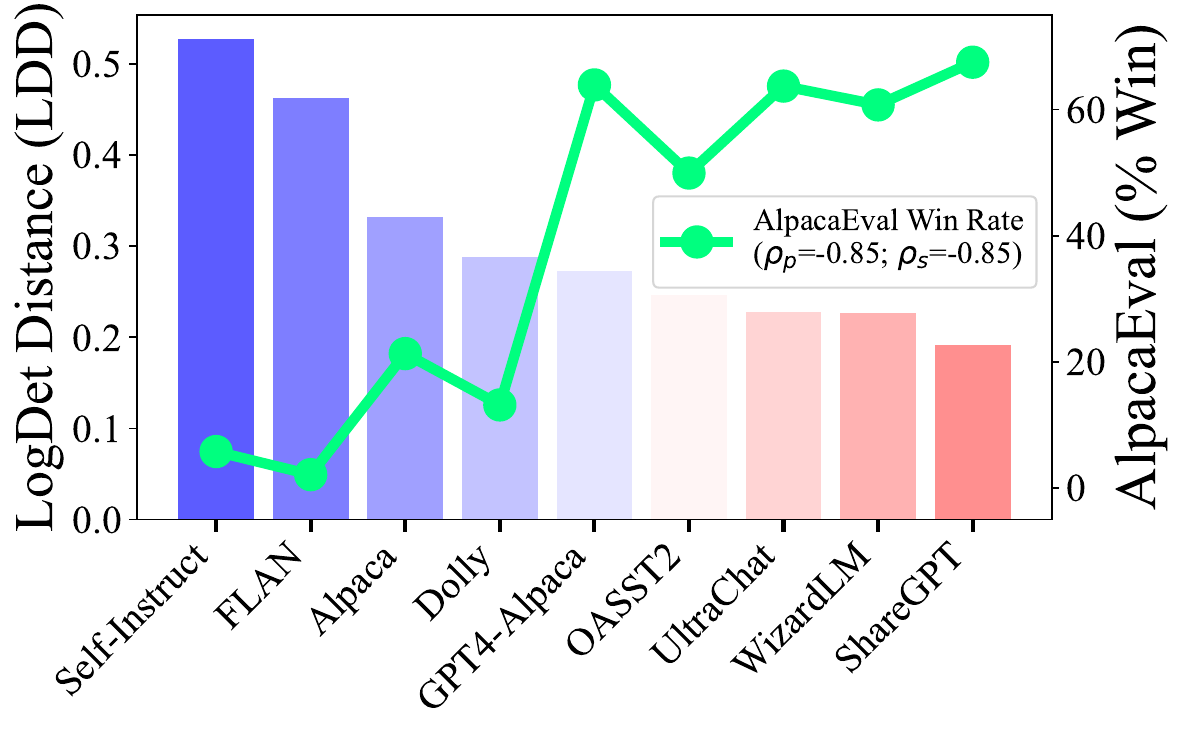}}
    \vskip -0.15in
    \caption{The log determinant distance in \cref{eq:ldd_definition} of instruction tuning datasets is correlated with instruction following performance when the model is finetuned on these datasets, with a Pearson correlation of $\rho_{p}=-0.85$ and a Spearman's rank correlation of $\rho_{s}=-0.85$.}
    \label{fig:fig_demo_diversity_vs_performance_across_dataset_LM_Weight_Gradient}
    \end{center}
    \vskip -0.35in
\end{figure}

\begin{figure*}[ht]
    \begin{center}
    \centerline{\includegraphics[width=\textwidth]{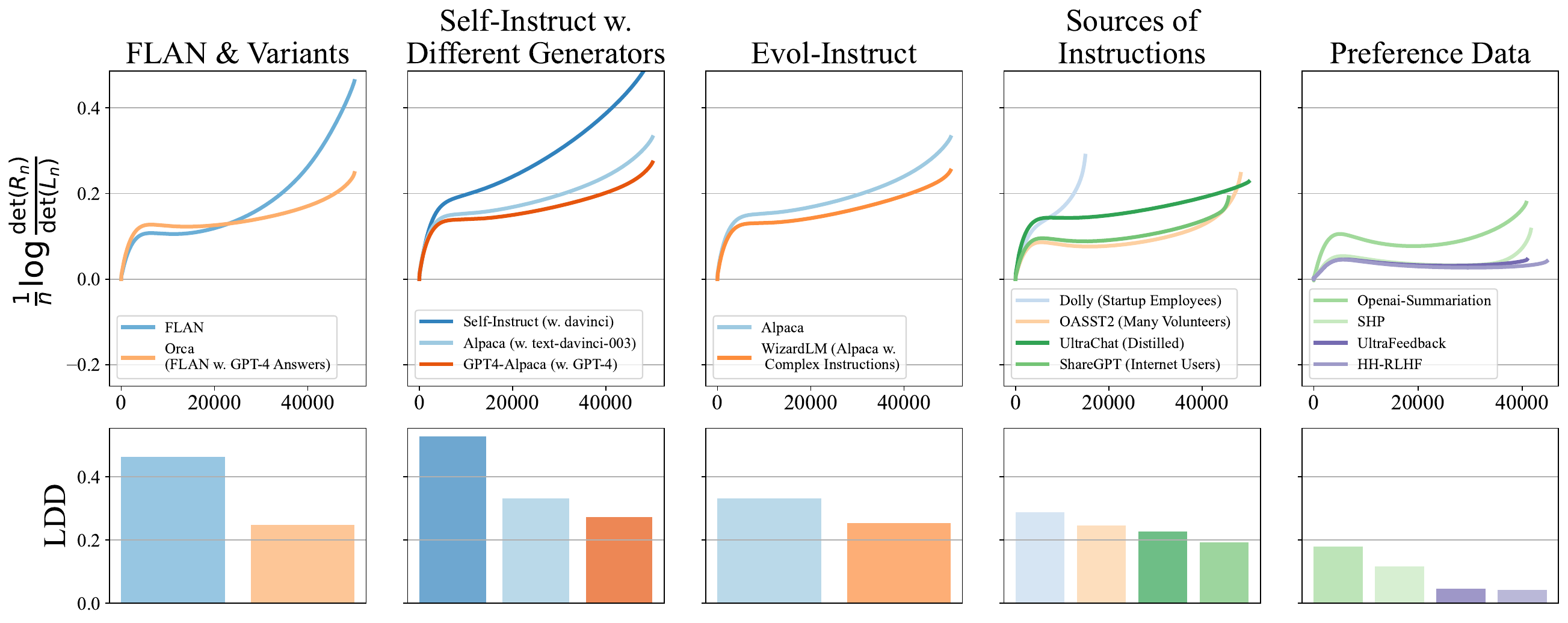}}
    \caption{Studies of the log determinant distance as a measure of diversity of instruction tuning and preference learning datasets. More diverse datasets yield a log determinant distance curve that is closer to a horizontal line and closer to zero. Distilling responses from capable large language models improves diversity (1st panel). The diversity of synthetic datasets generated using Self-Instruct \citep{wangSelfInstructAligningLanguage2023} increase with better teacher model (2nd panel). Using LLMs to re-write instructions to be more complex \citep{xuWizardLMEmpoweringLarge2024} also improves diversity (3rd panel). Curating instructions from diverse sources like ShareGPT \& OASST2 yields consistently higher average marginal gains compared to those curated with less human involvement (4th panel). Preference datasets overall are a lot more diverse than instruction tuning datasets, some with no apparent drop off in average marginal gains (5th panel).}
    \label{fig:fig_compare_dataset_groups_using_ldd_curves}
    \end{center}
    \vskip -0.25in
\end{figure*}

\subsection{Results: Diversity Assessment}

To assess the log determinant distance as a diversity measure, we compute the log determinant distance using weight gradient vectors $\nabla_{\theta}\ell$ on 9 instruction tuning datasets and 4 preference learning datasets detailed in \cref{sec:implementation_details}. For instruction tuning datasets, $\ell$ is the average log likelihood of tokens in the response conditioned on the instruction. For preference learning datasets, $\ell$ is the log odds of the preferred response over an alternative worse response.

The results \cref{fig:fig_demo_diversity_vs_performance_across_dataset_LM_Weight_Gradient} demonstrate that the log determinant distance of instruction tuning datasets is correlated with instruction following performance of models finetuned on these datasets. \cref{fig:fig_demo_how_to_compute_diversity_and_show_corr_with_performance_compate_data_representation} compares the log determinant distance of datasets computed across different data representations: \textsc{Mpnet Emb}, \textsc{Llama Emb}, and \textsc{Llama $\nabla_{\theta}\ell$}, and illustrates that \textsc{Llama $\nabla_{\theta}\ell$} is the only data representation that provides a useful predictor of instruction following performance.

\cref{fig:fig_compare_dataset_groups_using_ldd_curves} compares the log determinant distance of instruction tuning datasets and preference learning datasets. Using log determinant distance as a proxy for dataset diversity, there are a few takeaways: (1) the diversity of datasets improves from distilling responses or both instructions and responses from a performant LLM, (2) distilling from better teacher models improves dataset diversity even more, (3) rephrase instructions to be more complex also improves dataset diversity, (3) curating instructions from diverse sources, e.g., from real users on the internet or large-scale crowdsourcing, promotes dataset diversity, and (5) preference learning datasets are overall more diverse than instruction tuning datasets.

\subsection{Results: Data Selection with DPPs}
\label{sec:data_selection_with_dpps}

In this section, we benchmark our DPP data selection approach on two instruction tuning datasets of varying diversity: Alpaca \citep{alpaca2023} and UltraChat \citep{dingEnhancingChatLanguage2023}. The latter is more diverse than the former (\cref{fig:fig_demo_diversity_vs_performance_across_dataset_LM_Weight_Gradient}). The data budget is 20\% (10,000) of the total dataset size.

Baselines include random selection (\textsc{Random}), set-cover based deduplication (\textsc{Dedup}) \citep{abbasSemDeDupDataefficientLearning2023}. We also include several rank-and-select approaches based on the norm of the weight gradient (\textsc{$\norm{\nabla_{\theta}\ell}_2$}), ChatGPT ratings of examples (\textsc{Alpagasus Rating}) \citep{chenAlpaGasusTrainingBetter2024}, (\textsc{EL2N}) \citep{paulDeepLearningData2021}, instruction following difficulty (\textsc{IFD}) \citep{liQuantityQualityBoosting2023}, the perplexity of the response conditioned on the instruction (\textsc{Perplexity}), and token counts (\textsc{\#Input Tokens}, \textsc{\#Output Tokens}, \textsc{\#Total Tokens}). 

\cref{tab:pruning_results_on_two_datasets} reports the performance of models finetuned on data subsets selected using our method and baselines on \textsc{Benchmark Avg} for generic abilities and \textsc{AlpacaEval} for instruction following.  \cref{tab:pruning_results_on_two_datasets_detailed} provides further details.

We investigate the effect of data representation choice for our DPP-based approach. Using \textsc{Llama $\nabla_{\theta}\ell$} as the data representation yields the largest improvement in instruction following performance compared to alternative data representations, e.g., \textsc{Llama Emb} and \textsc{MpNet Emb}, on the Alpaca dataset. \cref{fig:fig_vmf_grad_vs_random_cross_datasets} illustrates that this is true for different data budgets. On the more diverse UltraChat dataset, random selection is a strong baseline, and different data representations exhibit similar performance.

We also assess data selection methods based on quality scores. In general, data selection with \textsc{EL2N} and \textsc{\#Input Tokens} results in subsets that perform worse than random subsets while using all other quality scores improve instruction following performance. Retaining examples with small $\norm{\nabla_{\theta}\ell}_2$, instead of large $\norm{\nabla_{\theta}\ell}_2$ typically used in active learning \citep{parkActiveLearningStrong2022}, leads to significant improvement. Selecting examples with large \textsc{\#Output Tokens} yields the most substantial improvement in instruction tuning performance, doubling the win rate compared to random selection on the Alpaca dataset and match the performance of finetuning on 100\% of the data on the UltraChat dataset.

\newcommand{\twodatasetsperftablecaption}{
    The performance of Llama-7b finetuned on 10k (20\%) data subset obtained using our DPP-based and alternative baseline data selection methods. We evaluate finetuned language models' generic abilities \textsc{Benchmark Avg} and instruction following abilities \textsc{AlpacaEval} on Alpaca and UltraChat. $(\uparrow)$ indicates that data points with higher quality scores are selected.
}

\begin{table*}[ht]
    \vskip -0.1in
    \caption{\twodatasetsperftablecaption}
    \label{tab:pruning_results_on_two_datasets}
    \centering\sc\small
    \begin{tabular}{l|ccr|ccr}
    \toprule
    Datasets & \multicolumn{3}{c}{Alpaca} & \multicolumn{3}{c}{UltraChat} \\
    Methods & \makecell{Benchmark \\ Avg} & \makecell{AlpacaEval \\ \% Win} & \makecell{ \\ Len} & \makecell{Benchmark \\ Avg} & \makecell{AlpacaEval \\ \% Win} & \makecell{ \\ Len} \\
    \midrule
    100\% Data & 23.0 & 21 & 91 & 22.9 & \textbf{64} & 215 \\
    Random & 21.6 & 18 & 90 & 22.6 & 57 & 213 \\
    \midrule
    DPP (Llama $\nabla_{\theta}\ell$) & 21.7 & 26 & 104 & \textbf{23.3} & 58 & 217 \\
    DPP (Llama $\nabla_{\theta}\ell$ Not Norm.) & 22.7 & 8 & 34 & 22.7 & 54 & 197 \\
    DPP (Llama Emb) & 22.5 & 21 & 92 & \textbf{23.3} & 53 & 204 \\
    DPP (Llama Emb Not Norm.) & 22.1 & 22 & 93 & 23.2 & 56 & 215 \\
    DPP (MpNet Emb) & 22.9 & 20 & 88 & \textbf{23.3} & 58 & 211 \\
    Dedup(MpNet Emb) & 22.8 & 19 & 85 & 23.0 & 58 & 219 \\
    \midrule
    $\norm{\nabla_{\theta} \ell}_2$ ($\downarrow$) & 22.7 & 37 & 136 & \textbf{23.3} & 63 & 249 \\
    Alpagasus Rating ($\uparrow$) & 21.6 & 21 & 95 & - & - & - \\
    EL2N ($\downarrow$) & 22.8 & 21 & 99 & 22.7 & 56 & 220 \\
    IFD ($\uparrow$) & 21.5 & 29 & 113 & 23.0 & 62 & 271 \\
    Perplexity ($\downarrow$) & 22.7 & 26 & 106 & 22.8 & 60 & 231 \\
    \#Input Tokens ($\uparrow$) & \textbf{25.1} & 19 & 88 & 22.9 & 51 & 217 \\
    \#Output Tokens ($\uparrow$) & 23.4 & 39 & \textbf{152} & 22.7 & \textbf{64} & 262 \\
    \#Total Tokens ($\uparrow$) & 20.4 & 38 & 149 & 22.7 & 62 & 251 \\
    \midrule 
    DPP (Llama $\nabla_{\theta}\ell$ + \#Output Toks) & 23.5 & \textbf{41} & 150 & 22.6 & \textbf{64} & \textbf{273} \\
    \bottomrule
    \end{tabular}
    \vskip -0.1in
\end{table*}

To investigate how DPP-based approach balances diversity and quality, we balance most effective quality score \textsc{\#Output Tokens} with diversity in the normalized weight gradient space. We assign a higher weight ($\lambda=0.9$) to \textsc{\#Output Tokens}. This approach leads to slightly improved instruction following performance compared to solely relying on \textsc{\#Output Tokens} on the Alpaca dataset and no improvement on the UltraChat dataset. \cref{fig:fig_dpp_weight_grad_and_numtoks_output_sweep_lambda} illustrates the diversity-quality trade-off of DPP-based selection method on a length adjusted win rate metric: AlpacaEval's win rate divided by the average output token lengths of the model's output, multiplied by that of the reference model's generations.

\section{Discussion}

We introduced a DPP-based approach to select instruction tuning data subsets that provide a flexible framework to integrate different notions of data similarity and quality. Our approach of measuring data similarity in the normalized gradient space improves instruction following performance over alternative data similarity measures on redundant dataset like Alpaca. More importantly, we proposed log determinant distance to quantify dataset diversity that is correlated with instruction following performance. We can use the proposed diversity metric to understand how much data should be kept when selecting data subsets. \cref{fig:fig_vmf_grad_vs_random_cross_datasets} illustrates that the less diverse a dataset is (e.g., Dolly and Alpaca), the more data could be pruned without sacrificing performance. This implies that we should adjust the data budget proportional to its diversity. Furthermore, it provides us with a way to compare the diversity of existing instruction tuning \& preference learning datasets to better understand the impact of the different curation strategies' on dataset diversity.

When assessing the performance of finetuned models using AlpacaEval \citep{duboisAlpacaFarmSimulationFramework2023}, we corroborated existing observations that GPT-4 judge favors more verbose responses. Notably, we found that selecting data subsets with long responses yielded the most substantial improvement in win rate compared to alternative data selection methods. One might argue that longer responses contain more information that users prefer. However, we'd want to understand as well as to optimize for the model's instruction following abilities after controlling for the length bias. Some work has started to address this issue of biased evaluation \citep{shenLooseLipsSink2023} that will be crucial for our problem of selecting optimal data subsets for instruction tuning.

Enforcing dataset diversity proves beneficial on less diverse datasets. This benefit diminishes when applied to more diverse datasets. In such cases, selecting datasets randomly after basic text deduplication may be adequate. Our research on dataset diversity is useful to determine the placement of a new dataset on the diversity spectrum, helping us understand whether it is worthwhile to implement more sophisticated ways to encourage diversity.

Our work suggests how to improve dataset diversity. We emphasize the importance of curating datasets with realistic instructions from diverse sources, e.g., internet user interactions with LLMs. If extensive human involvement is cost-prohibitive, an alternative approach is to distill the dataset entirely or re-write partially using the most capable LLMs. Surprisingly, preference learning datasets exhibit greater diversity compared to instruction tuning datasets, even if derived from the same source (e.g., UltraFeedback is curated from FLAN, UltraChat etc.). More work is required to better understand this phenomenon and its implications.

\section{Conclusion}

We present a DPP-based approach to select instruction tuning data subsets that prioritizes both diversity and quality. We proposes to use log determinant distance to measure dataset diversity that is useful for analyzing datasets and selecting data subsets.

\section{Impact}

This paper presents work whose goal is to advance the field of Machine Learning. There are many potential societal consequences of our work, none which we feel must be specifically highlighted here.

\newpage
\newpage

\bibliography{nlp_LM,dataset_selection,dpp,misc}
\bibliographystyle{icml2024}

\newpage
\appendix
\onecolumn

\section{Appendix}
\label{sec:appendixA}

\subsection{Proof of \cref{lemma:jl_lemma_rowwise_projection}}
\label{sec:appendixA_jl_lemma_rowwise_projection_proof}

Let $\epsilon,\delta$ be given. For notation convenience, let $p \equiv \vect(\nabla_W \ell)$ and $q \equiv \vect(\nabla_B \ell)$. Let $p_1, \cdots, p_m \in \R^n$ be rows of $\vect(\nabla_W\ell)$ and $q_1, \cdots, q_m \in \R^r$ be rows of $\vect(\nabla_B \ell)$. Due to \cref{eq:grad_B_is_projection_of_grad_W}, we have $q_k = A p_k$ for $k= 1, \cdots, m$. Provided $A\sim \sN(0,\frac{1}{r})$ and $r=\sO(\log(1/\delta)/\epsilon^2)$, the following holds due to Johnson-Lindenstrauss Lemma \citep{johnsonExtensionsLipschitzMappings1984}:
\begin{align}
    \Prob \pb{
        \left|
            \norm{q_k}_2^2 - \norm{p_k}_2^2
        \right|
        < \frac{\epsilon}{m}
    }
        \geq 1 - \frac{\delta}{m}.
\end{align}
By union bound,
\begin{align}
    \label{eq:union_bound_on_rows_of_grad}
    \Prob\pb{
        \bigcup_{k=1}^m \pc{
            \left|
                \norm{q_k}_2^2 - \norm{p_k}_2^2
            \right|
            > \frac{\epsilon}{m}
        }
    }
        \leq \sum_{k=1}^m \Prob\pb{
            \left|
                \norm{q_k}_2^2 - \norm{p_k}_2^2
            \right|
            > \frac{\epsilon}{m}
        }
        \leq \sum_{k=1}^m \frac{\delta}{m}
        \leq \delta.
\end{align}
If $\left|\norm{q_k}_2^2 - \norm{p_k}_2^2\right| < \frac{\epsilon}{m}$ for all $k=1,\cdots,m$, then
\begin{align}
    \left|
        \norm{q}_2^2 - \norm{p}_2^2
    \right|
        = \left| \sum_{k=1}^m \norm{q_k}_2^2 - \sum_{k=1}^m \norm{p_k}_2^2 \right|
        \leq \sum_{k=1}^m \left| \norm{q_k}_2^2 - \norm{p_k}_2^2 \right|
        \leq \sum_{k=1}^m \frac{\epsilon}{m}
        = \epsilon.
\end{align}
Therefore,
\begin{align}
    \Prob\pb{
        \left| \norm{q}_2^2 - \norm{p}_2^2 \right|
            \leq \epsilon
    }
        \geq \Prob\pb{
            \bigcap_{k=1}^m \pc{
                \left| \norm{p_k}_2^2 - \norm{q_k}_2^2 \right|
                    < \frac{\epsilon}{m}
            }
        }
        \geq 1 - \delta
\end{align}
where the last inequality is by \cref{eq:union_bound_on_rows_of_grad}.

\section{Appendix}
\label{sec:appendixB}

This section of the appendix contains additional tables and figures from \cref{sec:experiments}.

\begin{figure*}[ht]
    \begin{center}
    \centerline{\includegraphics[width=\textwidth]{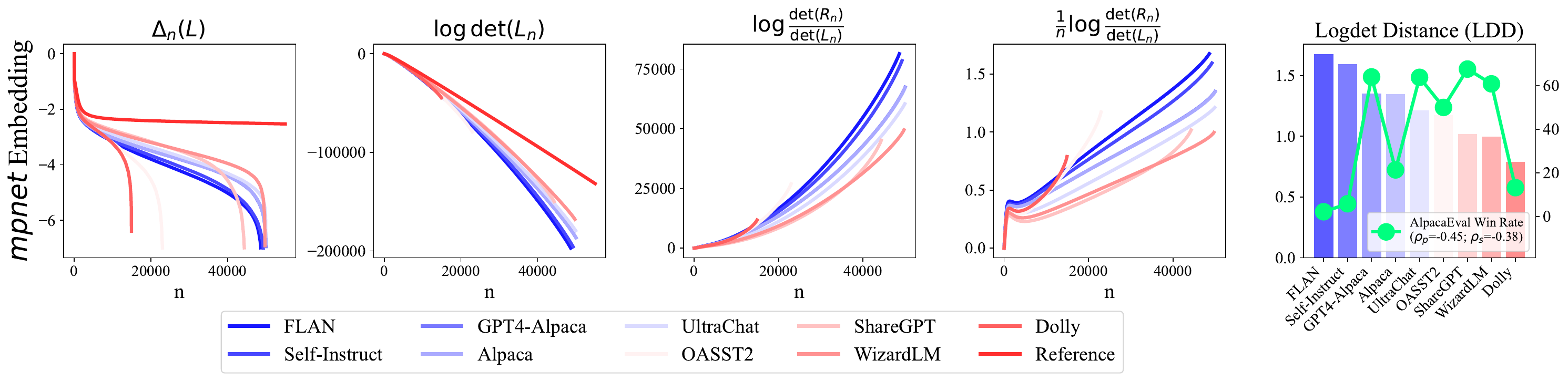}}
    \centerline{\includegraphics[width=\textwidth]{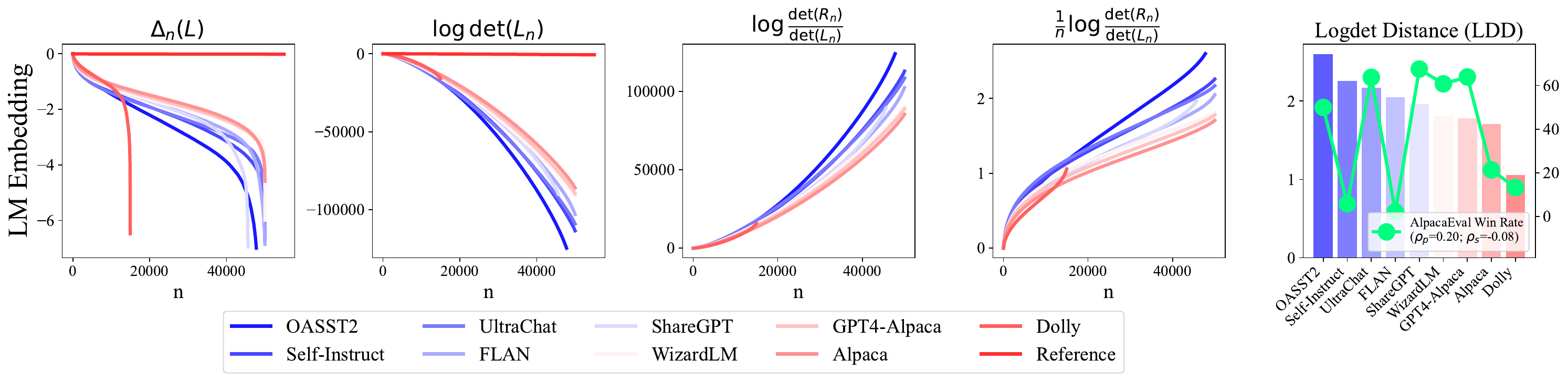}}
    \centerline{\includegraphics[width=\textwidth]{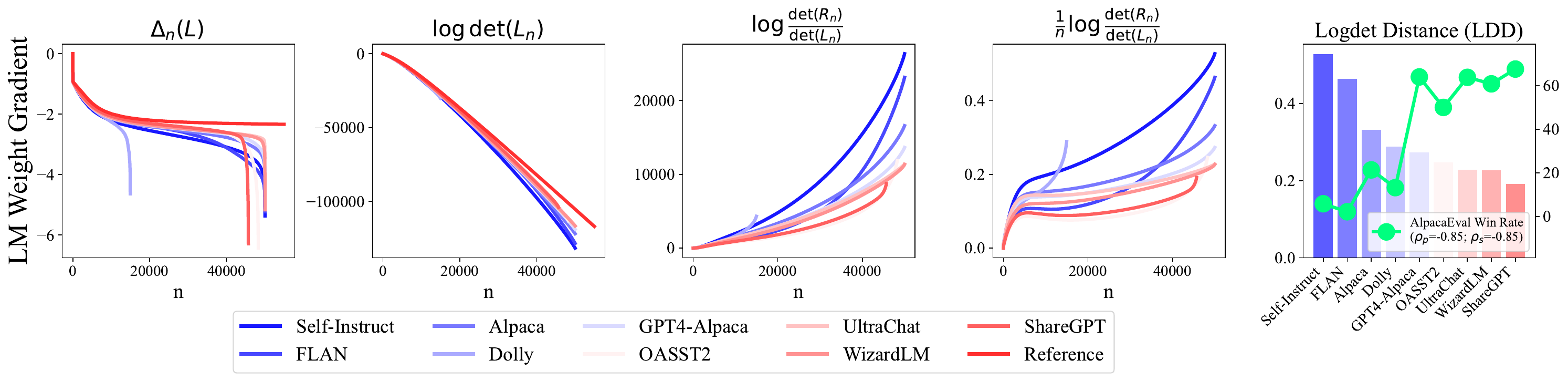}}
    \caption{Comparison of the log determinant distance computed using different data representations: \textsc{MpNet Emb} (top row), \textsc{Llama Emb} (middle row), and \textsc{Llama $\nabla_{\theta}\ell$} with respect to instruction tuning loss (bottom row). The log determinant distance based on weight gradient provides the strongest correlation with instruction following performance.}
    \label{fig:fig_demo_how_to_compute_diversity_and_show_corr_with_performance_compate_data_representation}
    \end{center}
\end{figure*}

\newcommand{\appendixalpacaperftablecaption}{%
    Academic benchmark and instruction following evaluation of Llama-7b finetuned on 10k (20\%) data subset of Alpaca (top block) and UltraChat (bottom block). This table compares random selection and full finetuning baseline as well as data selection methods that ensure diversity, quality, or both. $(\uparrow)$ indicates that data points with higher quality score are selected.
}

\begin{table*}[ht]
    \caption{\appendixalpacaperftablecaption}
    \centering\sc\small
    \begin{tabular}{l|cccccc|rr}
    \toprule
    Methods & \multicolumn{6}{c}{Academic Benchmarks} & \multicolumn{2}{c}{AlpacaEval} \\
     & MMLU & GSM & BBH & TydiQA & CodexEval & Avg & \% Win & Length \\
    \midrule 
    \multicolumn{9}{c}{Alpaca} \\ 
    \midrule
    Random & 32.3 & 4.8 & 33.4 & 22.4 & 8.5 & 21.6 & 18 & 90 \\
    100\% Data & 41.7 & 4.5 & 32.2 & 20.2 & 9.8 & 23.0 & 21 & 91 \\
    \midrule
    DPP (Llama $\nabla_{\theta}\ell$ Not Norm.) & 41.3 & 5.3 & 26.3 & \textbf{25.0} & 8.5 & 22.7 & 8 & 34 \\
    Dedup(MpNet Emb) & 38.0 & 5.8 & 32.1 & 21.4 & 11.0 & 22.8 & 19 & 85 \\
    DPP (MpNet Emb) & 39.1 & 5.8 & 33.7 & 20.0 & 8.5 & 22.9 & 20 & 88 \\
    DPP (Llama Emb) & 37.9 & 6.9 & 29.9 & 21.0 & 11.2 & 22.5 & 21 & 92 \\
    DPP (Llama Emb Not Norm.) & 36.9 & 5.1 & 31.6 & 21.5 & 8.7 & 22.1 & 22 & 93 \\
    DPP (Llama $\nabla_{\theta}\ell$) & 28.8 & \textbf{7.4} & 34.1 & 22.6 & 9.6 & 21.7 & 26 & 104 \\
    \midrule
    \#Input Tokens ($\uparrow$) & \textbf{43.8} & 5.3 & 34.0 & 24.8 & 10.4 & \textbf{25.1} & 19 & 88 \\
    EL2N ($\downarrow$) & 38.3 & 6.4 & 32.0 & 19.9 & \textbf{12.2} & 22.8 & 21 & 99 \\
    Alpagasus Rating ($\uparrow$) & 36.2 & 5.2 & 31.3 & 19.8 & 9.1 & 21.6 & 21 & 95 \\
    Perplexity ($\downarrow$) & 37.7 & 4.7 & 32.9 & 21.0 & 11.6 & 22.7 & 26 & 106 \\
    IFD ($\uparrow$) & 34.6 & 4.4 & 30.9 & 23.5 & 6.7 & 21.5 & 29 & 113 \\
    $\norm{\nabla_{\theta} \ell}_2$ ($\downarrow$) & 36.9 & 6.9 & 33.1 & 20.9 & 8.5 & 22.7 & 37 & 136 \\
    \#Total Tokens ($\uparrow$) & 25.1 & 7.2 & 33.5 & 21.7 & 8.5 & 20.4 & 38 & 149 \\
    \#Output Tokens ($\uparrow$) & 36.3 & 7.1 & \textbf{35.4} & 21.8 & 9.1 & 23.4 & 39 & \textbf{152} \\
    \midrule
    DPP (Llama $\nabla_{\theta}\ell$ + \#Output Toks) & 36.6 & 6.6 & 35.2 & 22.3 & 9.8 & 23.5 & \textbf{41} & 150 \\
    \midrule
    \multicolumn{9}{c}{UltraChat} \\ 
    \midrule
    Random & 36.2 & 7.5 & 32.8 & 19.8 & 11.0 & 22.6 & 57 & 213 \\
    100\% Data & 37.8 & 7.8 & 32.1 & 20.2 & 10.4 & 22.9 & \textbf{64} & 215 \\
    \midrule
    DPP (Llama Emb) & 38.1 & 8.9 & 32.8 & 18.9 & 12.2 & \textbf{23.3} & 53 & 204 \\
    DPP (Llama $\nabla_{\theta}\ell$ Not Norm.) & 37.7 & 8.4 & 32.6 & 17.8 & 11.0 & 22.7 & 54 & 197 \\
    DPP (Llama Emb Not Norm.) & 38.0 & 8.6 & 34.2 & 19.6 & 8.5 & 23.2 & 56 & 215 \\
    DPP (MpNet Emb) & 37.3 & 8.6 & 33.9 & 19.1 & 11.6 & \textbf{23.3} & 58 & 211 \\
    Dedup(MpNet Emb) & 37.4 & 8.7 & 31.5 & 21.2 & 9.1 & 23.0 & 58 & 219 \\
    DPP (Llama $\nabla_{\theta}\ell$) & 36.2 & 7.6 & \textbf{34.5} & 20.1 & \textbf{12.8} & \textbf{23.3} & 58 & 217 \\
    \midrule
    \#Input Tokens ($\uparrow$) & 37.7 & 8.5 & 33.6 & 18.5 & 9.1 & 22.9 & 51 & 217 \\
    EL2N ($\downarrow$) & \textbf{38.6} & 7.5 & 32.4 & 18.2 & 11.0 & 22.7 & 56 & 220 \\
    Perplexity ($\downarrow$) & 36.9 & 7.6 & 33.0 & 19.4 & 11.6 & 22.8 & 60 & 231 \\
    \#Total Tokens ($\uparrow$) & 34.6 & 8.3 & 33.3 & 20.9 & 9.8 & 22.7 & 62 & 251 \\
    IFD ($\uparrow$) & 34.5 & 10.2 & 33.4 & \textbf{21.4} & 7.9 & 23.0 & 62 & \textbf{271} \\
    $\norm{\nabla_{\theta} \ell}_2$ ($\downarrow$) & 34.7 & \textbf{10.3} & 32.2 & \textbf{21.4} & 12.2 & \textbf{23.3} & 63 & 249 \\
    \#Output Tokens ($\uparrow$) & 34.5 & \textbf{10.3} & 31.9 & 20.8 & 9.1 & 22.7 & \textbf{64} & 262 \\
    \midrule 
    DPP (Llama $\nabla_{\theta}\ell$ + \#Output Toks) & 34.5 & 8.9 & 32.3 & \textbf{21.6} & 9.1 & 22.6 & \textbf{64} & \textbf{273} \\
    \bottomrule
    \end{tabular}
    \label{tab:pruning_results_on_two_datasets_detailed}
    \end{table*}

\begin{figure}[ht]
    \vskip 0.1in
    \begin{center}
    \centerline{\includegraphics[width=.5\columnwidth]{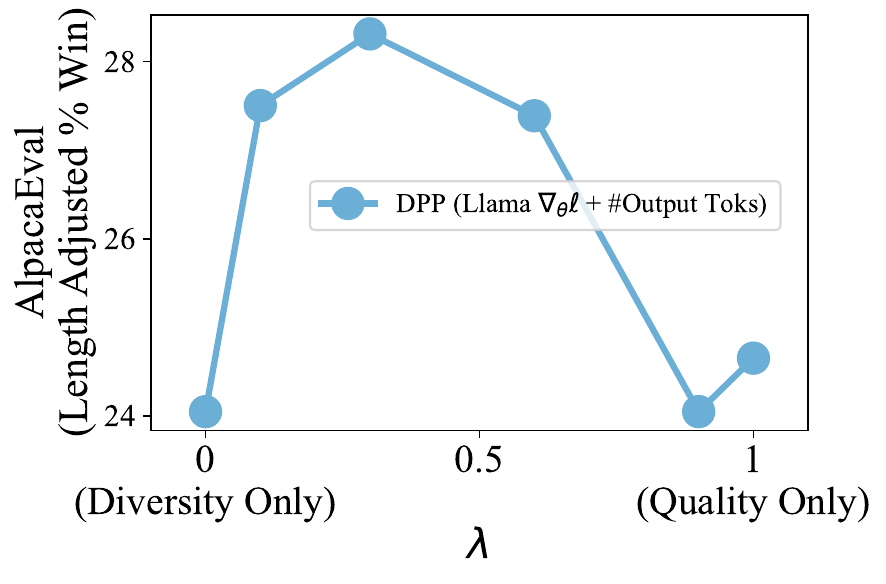}}
    \vskip -0.15in
    \caption{Vary $\lambda$ interpolates between enforcing diversity and selecting for quality. Here we use the length adjusted win rate metric computed from AlpacaEval's win rate divide by the average length of the model's generations, then multiply by that of the reference model's generations.}
    \label{fig:fig_dpp_weight_grad_and_numtoks_output_sweep_lambda}
    \end{center}
    \vskip -0.35in
\end{figure}

\begin{figure*}[ht]
    \begin{center}
    \centerline{\includegraphics[width=\textwidth]{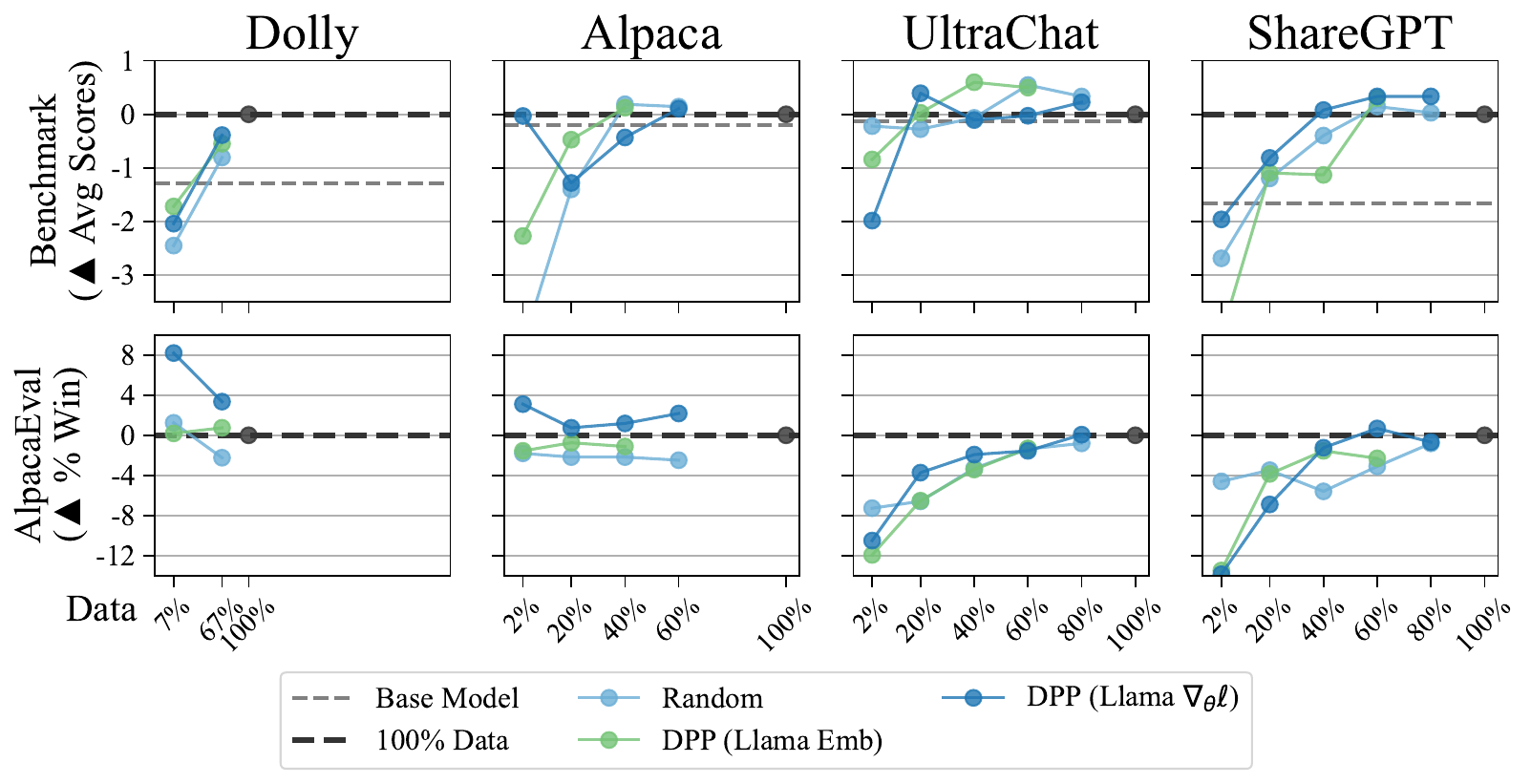}}
    \caption{Performance of DPP data selection methods relative to the full finetuning of Llama-7b on 4 datasets with different data budget. Using the \textsc{Llama $\nabla_{\theta}\ell$} as data representation is superior to \textsc{Llama Emb} on Dolly and Alpaca, while exhibiting comparable performance on UltraChat and ShareGPT. For (money) budget reasons, we use GPT-4-Turbo as the judge for AlpacaEval.
    }
    \label{fig:fig_vmf_grad_vs_random_cross_datasets}
    \end{center}
\end{figure*}

\end{document}